\documentclass[lettersize,journal]{IEEEtran}
\usepackage{amsmath,amsfonts}
\usepackage{algorithmic}
\usepackage{algorithm}
\usepackage{array}
\usepackage{amsmath}
\usepackage[caption=false,font=normalsize,labelfont=sf,textfont=sf]{subfig}
\usepackage{textcomp}
\usepackage{stfloats}
\usepackage{booktabs}
\usepackage{multirow}
\usepackage{xcolor}
\usepackage{url}
\usepackage{verbatim}
\usepackage{graphicx}
\usepackage{cite}
\usepackage{bigstrut}
\usepackage[pagebackref=true,breaklinks=true,colorlinks,bookmarks=false]{hyperref}

\usepackage{bbding}

\newcommand{\etal}{\textit{et al}.}
\newcommand{\ie}[1]{{\textit{i.e.}{{#1}}}}
\newcommand{\eg}[1]{{\textit{e.g.}{{#1}}}}

\begin{document}

\title{Efficient Token Compression for Vision Transformer with Spatial Information Preserved}

\author{Junzhu Mao, Yang Shen, Jinyang Guo, Yazhou Yao, and Xiansheng Hua

	\thanks{Junzhu Mao, Yang Shen and Yazhou Yao are with the School of Computer Science and Engineering, Nanjing University of Science and Technology, Nanjing 210094, China.} 
	\thanks{Jinyang Guo is with the State Key Laboratory of Software Development Environment, Institute of Artificial Intelligence, Beihang University, Beijing 100191, China} 
	\thanks{ Xiansheng Hua is with the Terminus Group, Beijing, 100027, China.} 
	\thanks{ Hengtao Shen is with the School of Computer Science and Engineering, University of Electronic Science and Technology of China, Chengdu 611731, China.}
	}

\markboth{IEEE Transactions on Multimedia}%
{Shell \MakeLowercase{\textit{et al.}}: A Sample Article Using IEEEtran.cls for IEEE Journals}

\maketitle

\begin{abstract}
 Token compression is essential for reducing the computational and memory requirements of transformer models, enabling their deployment in resource-constrained environments. In this work, we propose an efficient and hardware-compatible token compression method called Prune and Merge. Our approach integrates token pruning and merging operations within transformer models to achieve layer-wise token compression. By introducing trainable merge and reconstruct matrices and utilizing shortcut connections, we efficiently merge tokens while preserving important information and enabling the restoration of pruned tokens. Additionally, we introduce a novel gradient-weighted attention scoring mechanism that computes token importance scores during the training phase, eliminating the need for separate computations during inference and enhancing compression efficiency. We also leverage gradient information to capture the global impact of tokens and automatically identify optimal compression structures. Extensive experiments on the ImageNet-1k and ADE20K datasets validate the effectiveness of our approach, achieving significant speed-ups with minimal accuracy degradation compared to state-of-the-art methods. For instance, on DeiT-Small, we achieve a 1.64$\times$ speed-up with only a 0.2\% drop in accuracy on ImageNet-1k. Moreover, by compressing segmenter models and comparing with existing methods, we demonstrate the superior performance of our approach in terms of efficiency and effectiveness. Code and models have been made available at \url{https://github.com/NUST-Machine-Intelligence-Laboratory/prune_and_merge}.
\end{abstract}

\begin{IEEEkeywords}
Vision Transformer, Token Compression, Classification, Semantic Segmentation.
\end{IEEEkeywords}

\section{Introduction}

\IEEEPARstart{R}{ecently}, 
 Vision Transformer (ViT) has gained prominence within numerous computer vision tasks \cite{ZHANG2024102425,yao2021jo}, such as image recognition \cite{dosovitskiy2020image,SED,Jiang,sun2021webly}, object detection \cite{carion2020end,sun2024unified, yao2023automated,Cai}, semantic segmentation \cite{chen2025knowledge,yao2021non,chentaotip2024,10023953}, and video analysis \cite{pei2022eccv,tangcsvt,pei2024videomac,10105896,10298026}. Compared to Convolutional Neural Networks \cite{sheng2024adaptive,sun2022pnp}, the transformer architecture exhibits superior capacity for absorbing vast amounts of data and handling sequential inputs \cite{dosovitskiy2020image, touvron2021training, yuan2021tokens}. Consequently, researchers have been exploring the integration of transformers as a unified visual model architecture~\cite{chen2022unified, lu2022unified}, positioning transformers as a potential focus for the next generation of AI frameworks.

\begin{figure}[t]
	\centering
	\includegraphics[width=0.98\linewidth]{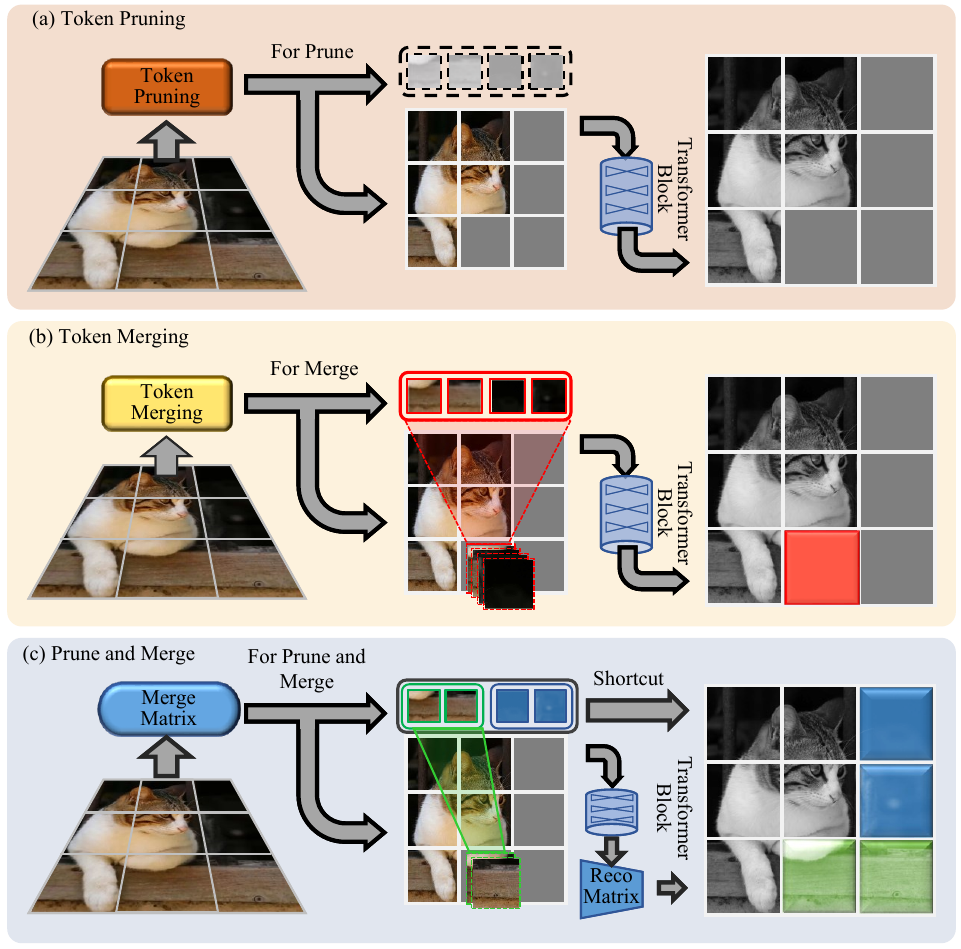}
	\caption{Our paper compares three token compression paradigms: (a) the regular token pruning, (b) the regular token merging operation, and (c) our Prune and Merge compression method. In contrast to (a) and (b), our proposed method seamlessly integrates token pruning and merging operations. Additionally, we reconstruct tokens in the transformer block output using a reconstruct matrix (labeled as "Reco Matrix" in the figure) and shortcut connections, enabling efficient layer-wise compression.}
	\label{fig1}
\end{figure}

However, the sequence-to-sequence nature of transformers presents computational efficiency challenges due to the quadratic scaling of computation cost with an increase in input tokens. This limitation hinders the deployment of ViT models on resource-limited devices like mobile phones or embedded systems. In light of this, researchers have turned to pruning, a widely studied model compression technique within CNNs~\cite{fan2019reducing, ganesh2021compressing, guo2019reweighted, michel2019sixteen, mccarley2019structured, chen2020lottery}known to extend effectively to transformer model compression~\cite{chen2021chasing, yu2021unified, yu2021aunified, zhu2021visual}.
Since the computational cost of transformers is primarily influenced by the number of tokens, as discussed in Sec~\ref{sec3.1}, token pruning becomes a promising subfield of ViT that can effectively accelerate models~\cite{goyal2020power, pan2021ia, pan2021scalable, rao2021dynamicvit, tang2022patch, xu2022evo, kong2021spvit, liang2021evit, zong2021self}.

Nevertheless, token pruning is fraught with issues. As depicted in Fig~\ref{fig1} (a), token pruning involves the removal of tokens within the transformer block, leading to a loss of information and restricting the number of prunable tokens~\cite{bolya2023token}. Furthermore, this process disrupts the spatial structure, requiring a feasible method to restore the positional information of pruned tokens, which can be challenging when adapting to downstream tasks~\cite{chang2023making}.
To overcome these obstacles, some researchers propose token merging as a solution~\cite{bolya2023token, liang2022expediting, chang2023making}, illustrated in Fig~\ref{fig1} (b), where tokens are combined instead of being pruned. This approach helps mitigate information loss incurred during the reduction of tokens and facilitates the restoration of the original spatial structure. 

Despite notable advancements in model compression through token merging, several challenges still require attention. Firstly, completely disregarding token pruning could result in lost benefits in terms of noise reduction and improved model generalization~\cite{lecun1990optimal, molchanov2017variational, han2016dsd, bartoldson2020generalization, keskar2016large}. In some cases, pruning can be more effective than merging, especially when compressing on a smaller scale.
Secondly, existing merging methods typically adopt a uniform or gradual approach without accounting for varying attention distributions across different layers~\cite{chang2023making}. This coarse-grained compression approach can lead to unsatisfactory performance. Therefore, a layer-wise method is needed, requiring the integration of compression modules at each layer. This brings us to the third challenge: the demand for high-efficiency compression methods.
However, current mainstream compression techniques, such as importance scores calculation based on attention maps~\cite{xu2022evo, kong2021spvit, liang2021evit}, iterative token clustering~\cite{liang2022expediting}, and bipartite matching for token pairing~\cite{bolya2023token}, often lack efficiency and heavily rely on specialized hardware support, making them less suitable for edge devices.

In response to these challenges, we propose the Prune and Merge method: an efficient, hardware-friendly compression technique that attains layer-wise compression while retaining the benefits of both token pruning and merging (see Fig~\ref{fig1} (c)).
Our method introduces a Prune and Merge module, which includes a learnable merge matrix and a reconstruct matrix, enabling token merging and pruning operations simultaneously. The addition of shortcut connections ensures efficient restoration of pruned tokens, while the reconstruct matrix aids in recovering token structure and positional information. Notably, our module uses basic matrix operations, making it highly compatible with various hardware types.
To further enhance the efficiency of our compression method, we incorporate a gradient-weighted attention scoring mechanism. This mechanism learns the dataset prior distribution through gradient-weighted attention and computes token importance scores during the training process. Consequently, there is no need for separate token importance calculations during inference, resulting in a significant improvement in overall compression efficiency.
Based on the computed importance scores, we categorize tokens into three groups to generate initial values for the merge matrix. High-scoring tokens are preserved, low-scoring tokens are pruned, and medium-scoring tokens are merged with nearby tokens of the same category. This classification-based approach facilitates effective token pruning and merging operations based on token importance.
Furthermore, We also employ gradient information to capture the global impact of tokens, guiding our automatic global token compression approach and facilitating the identification of winning tickets~\cite{chen2020lottery}, which represent optimal compression structures. The main contributions of this work are as follows:

(1) We present a novel token compression module that combines token pruning and merging within transformer models. By introducing a learnable merge matrix and leveraging shortcut connections, our method achieves layer-wise token compression, effectively reducing computational costs while preserving crucial information. Compared to existing methods, our approach offers a more efficient and hardware-friendly solution for token compression.

(2) We propose a gradient-weighted attention scoring mechanism that leverages the training process to learn the prior distribution of the dataset and compute token importance scores. Incorporating this approach eliminates the need for separate token importance calculations during inference, significantly improving the overall efficiency of model compression.

(3) We introduce an automatic global token compression approach that utilizes gradient information to capture the global impact of tokens. By leveraging this information, our method identifies winning tickets. Extensive experiments conducted on various benchmarks and ViT architectures demonstrate the superiority of our proposed Prune and Merge method.

\section{Related Works}
\label{related_work}
In this section, we briefly review the most related works of efficient transformer and transformer compression.

\subsection{Efficient Vision Transformers.}
Dosovitskiy et al. (2020) introduced vision transformers (ViT), which integrated pure transformers into the network architecture for visual tasks. ViT divides images into patches, converts them into token embeddings, and feeds them into transformer encoders. With ample training data, ViT surpassed convolutional neural networks (CNN) in various vision tasks. Subsequently, several variants of ViT have been proposed.
DeiT~\cite{touvron2021training} and T2T-ViT~\cite{yuan2021tokens} aimed to enhance ViT's training data efficiency through techniques such as teacher-student training and improved architectures.
LeViT~\cite{graham2021levit} and PVT~\cite{wang2021pyramid} introduced deep-narrow transformer structures that reduced the number of tokens in intermediate layers by incorporating sub-sampling operations like local average pooling and convolutional sampling.
MViT~\cite{fan2021multiscale} incorporated multiscale feature hierarchies with transformer models to capture different levels of visual information, while MViTv2~\cite{li2022mvitv2} further enhanced this approach with decomposed relative positional embeddings and residual pooling connections.
STViT~\cite{chang2023making} designed a semantic token vision transformer that generated semantic tokens with high-level information representation to replace redundant tokens.
In contrast to these methods that focused on designing efficient attention structures with additional parameters requiring learning from scratch, our work is centered around leveraging pre-trained models and adapting them to hardware requirements through token compression, requiring only a few epochs of training.

\begin{figure*}[h]
	\centering
	\includegraphics[width=1.0\linewidth]{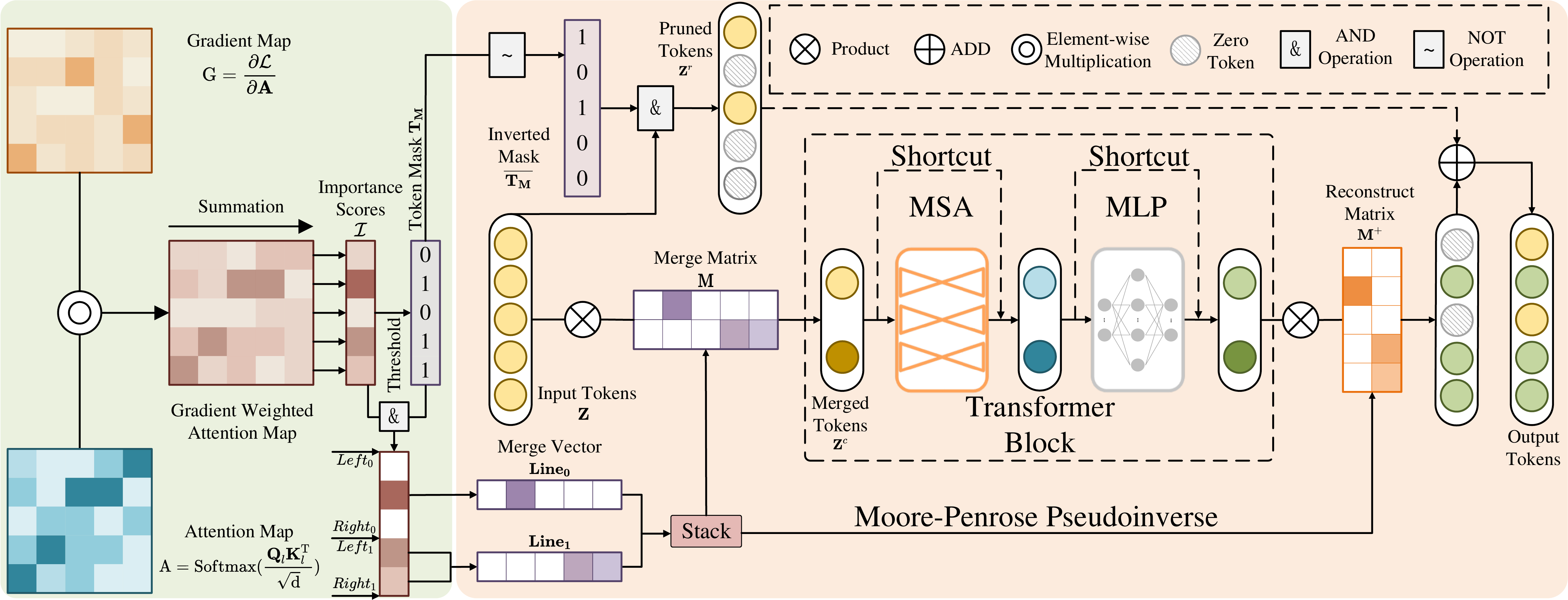}
	\caption{The architecture of our proposed approach.The left part illustrates the calculation process of the importance scores. We utilize the attention map $\mathrm{A}= \mathrm{Softmax}(\frac{\mathbf{Q}\mathbf{K}^{\mathrm{T}}} {\sqrt{\mathrm{d}}})$, which reflects the model's focus on different regions of the input, and its corresponding gradient map $ \mathrm{G}= \frac{\partial \mathcal{L}}{\partial \mathbf{A}} $, which shows the impact of each pixel on the prediction, to obtain the gradient-weighted map.  By summing along the Key dimension, we derive the importance scores. Based on the importance scores, we generate the token mask by applying a threshold.  The token mask allows us to zero out the scores of pruned tokens.  Moreover, employing specific algorithms, we derive both the merge matrix and the reconstruct matrix using the importance scores.The right part of the figure showcases the structural overview of our prune and merge method.  Initially, the input tokens undergo matrix multiplication with the merge matrix, resulting in compressed tokens.  These compressed tokens are then fed into the transformer block.  To restore the spatial resolution, the output of the block is multiplied by the reconstruct matrix. The pruned tokens are preserved using the negation of the token mask.  By incorporating shortcut connections, the pruned tokens are added to the output results, ensuring the retention of feature information.}
	\label{fig_framework}
\end{figure*}

\subsection{Prunning for ViT.} 
ViT pruning encompasses two primary methodologies: token pruning and weight pruning.  Weight pruning focuses on reducing specific structures' dimensions, such as attention heads~\cite{chen2021chasing, mao2023attention}, embedding dimensions~\cite{zhu2021visual}, hidden layer dimensions of the MLP~\cite{yu2021unified}, or the unified structure of the ViT model~\cite{yu2021unified}.Some works employ more distinctive techniques. For example, SAViT~\cite{zheng2022savit} integrates structural-aware interactions among components to enable collaborative pruning, and SPViT~\cite{he2024spvit} streamlines multi-head self-attention layers into convolutional operations within a single search path.  Despite weight pruning proves effective in reducing parameters and the previous works have achieved remarkable accomplishments, it may not efficiently address the substantial computational demands associated with high-resolution input images.

Token pruning, on the other hand, is specifically designed to address this computation problem. Various token pruning methods have been proposed to dynamically and hierarchically drop visual tokens at different levels using a learned token selector~\cite{pan2021ia}, an MLP-based prediction module for estimating importance scores~\cite{rao2021dynamicvit}, or employing max-pooling hierarchically~\cite{pan2021scalable}. Some methods even use reconstruction error of the pre-trained model to remove redundant patches based on a top-down patch slimming approach~\cite{tang2022patch}. Recently, Top-K pruning has been highlighted as a strong baseline for various classification tasks~\cite{haurum2023tokens}.
However, these token pruning methods often suffer from indiscriminate token removal, leading to unintended information loss and suboptimal performance, particularly at high pruning rates. To address these challenges while retaining the benefits of token pruning, our method incorporates token merging alongside token pruning.

\subsection{Token Merging.} 
Token merging is proposed as a solution to mitigate the information loss caused by token pruning~\cite{cao2023pumer}. Researchers in previous studies~\cite{xu2022evo, liang2021evit, kong2021spvit, zong2021self} have addressed token information preservation by merging the unimportant tokens. Xu~\etal~\cite{xu2022evo} proposes structure preserving token selection and slow-fast updating strategies to preserve the image information during pruning. Liang~\etal~\cite{liang2021evit} and Kong~\etal~\cite{kong2021spvit} fuse the less informative tokens to a new token for less information loss. Zong~\etal~\cite{zong2021self} propose a generic self-slimmed learning method that softly integrates redundant tokens into fewer informative ones. These methods combine unimportant tokens without considering the preservation of the spatial structure and position information for downstream input. ToMe~\cite{bolya2023token} gradually combines similar tokens using a bipartite soft matching algorithm. Expediting~\cite{liang2022expediting} uses an iterative clustering algorithm to get the cluster center of neighboring tokens. TPS~\cite{wei2023joint} squeezes the information of pruned tokens into partially reserved tokens via the unidirectional nearest-neighbor matching and similarity-based fusing steps. These methods merge similar or neighboring tokens to make the compression more accurate and restore spatial information easier. However, these methods, such as iterative clustering and similarity matching, are not highly efficient and often require hardware support, making them less suitable for edge devices. In contrast, our approach achieves token pruning and merging through matrix multiplication, enabling efficient and hardware-friendly layer-wise token compression. By combining the advantages of pruning to prevent overfitting, our method achieves efficient token reduction while maintaining competitive performance.

\section{The Proposed Approach}

In this section, we first introduce the preliminaries of ViT and show the pruning efficiency of token pruning by complexity analysis. Then our proposed Prune and Merge module and gradient-weighted attention scoring mechanism will be discussed in detail. Lastly, we further design a layer-weighted global pruning method to find the original model's winning ticket (\ie, the most suitable structure) automatically.

\subsection{Preliminaries and Complexity Analysis}
\label{sec3.1}

Vision Transformer~\cite{dosovitskiy2020image} does not change the standard transformer’s structure and accepts the one-dimensional token embedding sequence as input. According to this design, ViT reshapes the input image into flattened sequential patches and linearly projects them into embedding vectors. It also concatenates a trainable class token to the head of the patch sequence and adds a two-dimensional position embedding. The main body of ViT is stacked with serval transformer blocks, and the block can be divided into multi-head self-attention (MSA) and multi-layer perceptron (MLP) parts. 

\begin{table*}[t]
	\renewcommand\arraystretch{1.1}
	\centering
	\caption{The computational cost of each operation in a ViT block. $N$ and $D$ indicates the number of tokens and embedding dimension in ViT, respectively.}
		\begin{tabular}{c|c|c|c|c|c}
			\toprule
			Module  &  Input Size &     Operation        &  Weight Size         &      Output Size      & Computation     \\
			\hline
			\multirow{4}{*}{SA}  &
			$ 3\times N \times D$ & $QKV$ Projection     & $3\times D\times D$  & $ 3\times N \times D$ &   $3ND^2$      \bigstrut[t] \\ 
			&$ N \times D$        & $Q$ Multiplying $K^T$&         -            & $ N \times N$         &   $N^{2}D $     \\
			&$ N \times N$        & Multiplying $V$      &         -            & $ N \times D$         &   $N^{2}D $     \\
			&$ N \times D$        & Out Projection       &    $D\times D$       & $ N \times D$         &   $ND^2$        \\
			\hline
			\multirow{2}{*}{MLP}  &
			$ N \times D$         & FC Layer             &    $D\times 4D$      & $ N \times 4D$        &   $4ND^2$       \bigstrut[t]\\
			&$ N \times 4D$       & FC Layer             &    $4D\times D$      & $ N \times D$         &   $4ND^2$       \\
			\hline
			\multicolumn{5}{c}{Total Computational Complexity}                  & $ 12ND^{2} + 2N^{2}D $		          \bigstrut[t]\\
			\bottomrule	
		\end{tabular}
		
		\label{tab_analyze}
	\end{table*}  

For the input $\mathbf{Z}^{l-1}\in \mathbb{R}^{N\times D}$ of layer $l$ with $N$ tokens and embedding dimension $D$, the calculation of self-attention (SA) is as follows:
\begin{equation}
\mathrm{SA} (\mathbf{Z}_{l-1}) = \mathrm{Softmax}(\frac{\mathbf{Q}_{l} \mathbf{K}_{l}^{\mathrm{T}}} {\sqrt{\mathrm{d}}}) \mathbf{V}_l\mathbf{W}_l^\mathrm{o}.
\label{eq_1}
\end{equation}
Where Q, K, and V indicate the query, key, and value generated by three different input projections, $\mathbf{Q}_{l} = \mathbf{Z}_{l-1}\mathbf{W}_l^{\mathrm{q}}$, $\mathbf{K}_{l}=\mathbf{Z}_{l-1}\mathbf{W}_l^{\mathrm{k}}$, and $\mathbf{V}_l=\mathbf{Z}_{l-1}\mathbf{W}_l^{\mathrm{v}}$. $\mathbf{W}_l^\mathrm{o}$ is the output projection matrix after self-attention and $\mathrm{d}$ is the head embedding dimension. For simplicity and to maintain focus, we only discuss single-head attention in our analysis, as the multi-head case has the same computational complexity.
The MLP layer consists of two fully connected (FC) layers, which can be formulated as follows:
\begin{equation}
\mathrm{MLP}(\mathbf{Z}^{'}_{l}) = \mathrm{\phi} \left ( \mathbf{Z}_{l}^{'} \mathbf{W}_{l}^{\mathrm{fc1}} \right) \mathbf{W}_{l}^{\mathrm{fc2}}.
\label{eq_2}
\end{equation}
$\mathbf{Z}^{'}_{l}$ indicates the input of MLP and $\mathbf{W}_{l}^{\mathrm{fc1}}$ , $\mathbf{W}_{l}^{\mathrm{fc2}}$ are the weights of two FC layers. We use $\phi(\cdot)$ to represent the non-linear activation function (\eg, GeLU). Combining Eq.~\eqref{eq_1} and Eq.~\eqref{eq_2}, the block $\mathcal{B}_l (\cdot)$ can be defined as:
\begin{equation}
\mathcal{B}_{l} (\mathbf{Z}_{l-1}) = \mathrm{MLP}(\mathrm{SA}(\mathbf{Z}_{l-1}) + \mathbf{Z}_{l-1}) + \mathbf{Z}_{l}^{'}.
\label{eq_3}
\end{equation}

We show the detailed operation of each process and analyze its computation cost in Table~\ref{tab_analyze}. We only provide the single-head case to simplify the representation for the aforementioned reasons. $N$ represents the number of tokens, and $D$ represents the embedding dimension. They are the critical numbers of token pruning and parameter pruning focus on, respectively. From the final complexity formula, $N$ and $D$ contribute equally to the complexity. However, $D$ is a fixed value determined when the model is built, while $N$ will grow in square order with the growth of the input image size. Thus, $N$ can dominate the final computation, leading to a $O(N^{2}D)$ computation complexity. In this case, it is evident that token pruning, which reduces the number of $N$, can reduce the computation more efficiently than parameter pruning.

\subsection{Prune and Merge Module for Token Compression}

\label{sec3.2}

We propose the Prune and Merge module, as illustrated on the right side of Fig.~\ref{fig_framework}. Our module achieves token compression while preserving the spatial structure and positional information. By performing a matrix multiplication operation between the merge matrix and the input tokens, our module effectively combines token pruning and merging objectives. The output tokens of the transformer block are then obtained in the compressed format. To restore the original token count, we utilize a recovery matrix, which reverts the output tokens.

Starting with the original input tokens $\mathbf{Z}_{l-1} \in \mathbb{R}^{N \times D}$, where $N$ is the token count and $D$ signifies the embedding dimension. We initiate the compression by multiplying the merge matrix $\mathbf{M} \in \mathbb{R}^{M \times N}$, thereby reducing the number of tokens to $M$. The compression formula is as follows:
\begin{equation}
\mathbf{Z}_{l-1}^{c} \in \mathbb{R}^{M \times D} = \mathbf{M} \in \mathbb{R}^{M \times N} \times \mathbf{Z}_{l-1} \in \mathbb{R}^{N \times D}.
\label{eq_4}
\end{equation}
This operation yields the compressed input tokens $\mathbf{Z}_{l-1}^{c} \in \mathbb{R}^{M \times D}$. Subsequently, we preserve the pruned tokens $\mathbf{Z}_{l-1}^{r} \in \mathbb{R}^{n \times d}$ by performing a logical AND operation with the input tokens $\mathbf{Z}_{l-1}$ and the inverted token mask $\overline{\mathbf{T}_\mathbf{M}} \in \mathbb{R}^{N}$:
\begin{equation}
\mathbf{Z}_{l-1}^{r} \in \mathbb{R}^{N \times D} = 
\mathbf{Z}_{l-1} \in \mathbb{R}^{N \times D}~ \&~ \overline{\mathbf{T}_\mathbf{M}} \in \mathbb{R}^{N}.
\label{eq_5}
\end{equation}
The pruned tokens are retained, while others are set to zero. Finally, to recover the original dimension, we multiply the output of the transformer block with the Moore-Penrose pseudoinverse of the merge matrix $\mathbf{M^+} \in \mathbb{R}^{N \times M}$. By adding $\mathbf{Z}_{l-1}^{r}$ to the output tokens via shortcut connections, we ensure the retention of their feature information. The final output $\mathbf{Z}_{l} \in \mathbb{R}^{N\times D}$ is formulated as follows:
\begin{equation}
\mathbf{Z}_{l} = \mathbf{M^+} \in \mathbb{R}^{N \times M} \times
\mathcal{B}_{l} (\mathbf{Z}_{l-1}^{c}) \in \mathbb{R}^{M \times D} + \mathbf{Z}_{l-1}^{r}.
\label{eq_7}
\end{equation}

Our Prune and Merge module efficiently achieves both token pruning and merging through the use of a merge matrix. By merging similar tokens into a single token, we effectively achieve token compression while mitigating the information loss caused by the reduction in token count. Token pruning, on the other hand, is accomplished by setting the corresponding weights of the merge matrix to zero for tokens deemed least important. Despite the potential disadvantages brought about by token pruning, we believe that the removal of some unimportant tokens can have positive consequences on attention calculations and lead to an increase in model accuracy. Previous researches verifies that the removal of a subset of tokens can enhance the model's generalization ability and help prevent overfitting~\cite{lecun1990optimal, molchanov2017variational, han2016dsd, bartoldson2020generalization, keskar2016large}.
In our subsequent experiments, we will empirically verify the effectiveness of improving model accuracy through the targeted pruning of a small number of tokens.

\begin{algorithm}[t]
	\caption{Merge Matrix Generation}
	\begin{algorithmic}[1] 
		\REQUIRE 
		Importance scores $\mathcal{I}$ of $n$ tokens, PM-Threshold $\tau$, number of important tokens $m$, left and right indices of tokens to merge $Left, Right$, merge vector $\mathbf{Line} \in \mathbb{R}^{n}$, merge matrix $\mathbf{M} \in \mathbb{R}^{m \times n}$;
		
		\STATE Sort $\mathcal{I}$ in descending order to obtain $\mathbf{S}$;
		\STATE Determine the pruning threshold $\tau_1$ based on $\tau$ and $\mathbf{S}$;
		\STATE Set all elements of $\mathcal{I}$ that are less than $\tau_1$ to zero;
		\STATE Let $\tau_2$ be the $m$-th largest element in $\mathbf{S}$;
		\STATE Obtain indices $\mathbf{I}$ of elements in $\mathcal{I}$ that are greater than $\tau_2$;
		
		\STATE $Left \gets 1$;
		\FOR{each $i \in [1,m]$}
		\STATE $Right \gets \mathbf{I}[i]$;
		\IF{$Left < Right$}
		\STATE Set $\mathbf{Line}$ to the zero vector;
		\STATE $\mathbf{Line}[Left:Right] \gets \mathrm{Norm}(\mathcal{I}[Left:Right])$;
		\STATE $\mathbf{M}[i] \gets \mathbf{Line}$;
		\STATE $Left \gets Right$;
		\ENDIF
		\ENDFOR
		
		\ENSURE 
		Merge matrix $\mathbf{M}$; 
	\end{algorithmic}
	\label{alg_1}
\end{algorithm}

\begin{figure}[t]
	\includegraphics[width=0.98\linewidth]{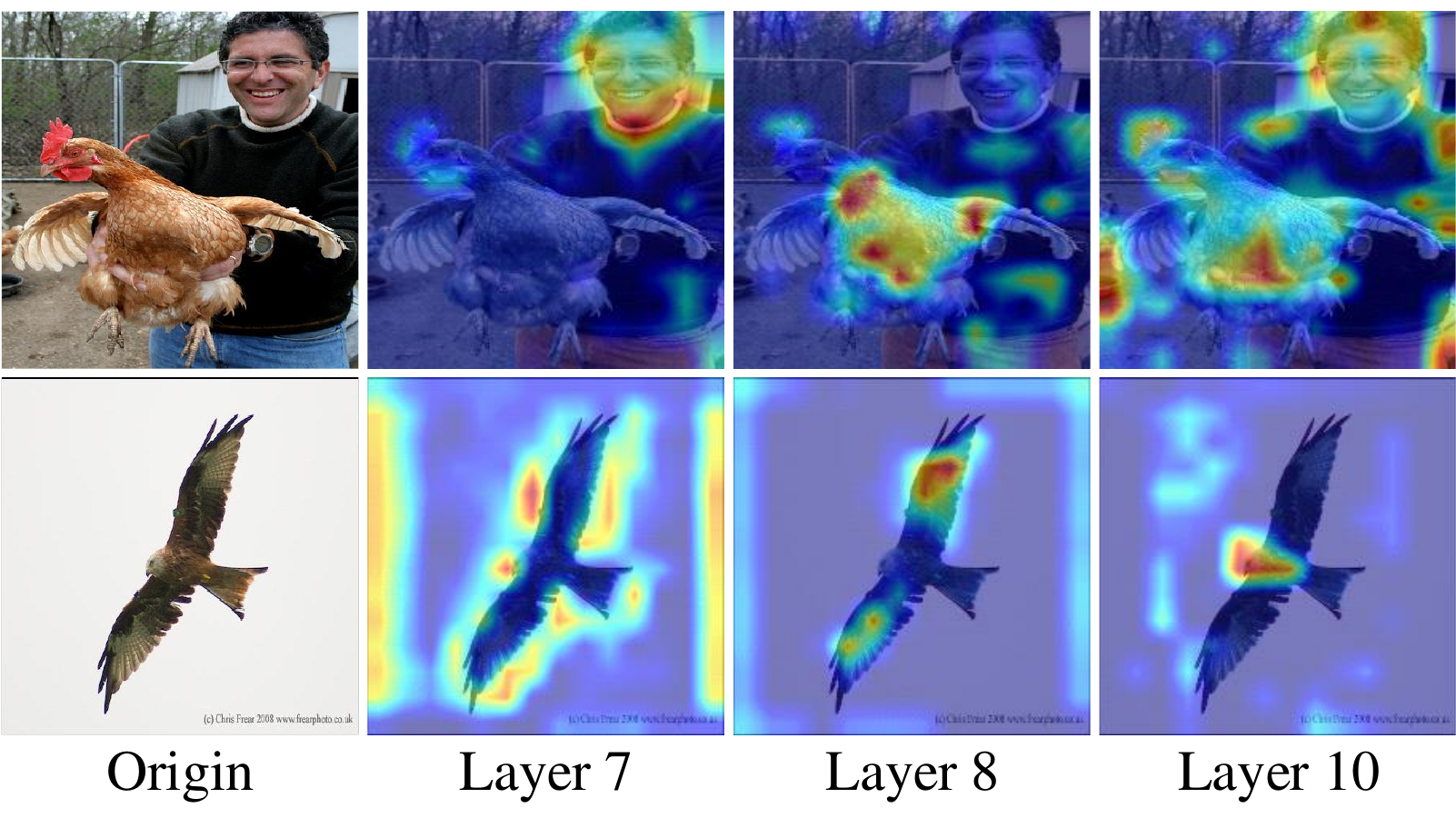}
	\vspace{-0.2cm}
	\caption{Grad-CAM visualization of ViT-Base's attention map in different layers. The attention on input images varies on different layers.}
	\label{fig_cam}
\end{figure}

Our Prune and Merge module achieves layer-wise token compression by employing a reconstruction matrix and shortcut connections, which effectively restore compressed tokens. This approach offers two key benefits.
Firstly, each block in our method utilizes a distinct pruning and merging scheme. As depicted in Fig.\ref{fig_cam}, visualized using Grad-CAM\cite{sel2017grad}, the token attention across different layers of the ViT can vary significantly. Some layers may even focus on background regions. We argue that ViT enhances its overall performance by learning from the context surrounding the targets. Consequently, adopting the same fusion scheme for all layers is inappropriate and may lead to decreased accuracy.
Secondly, our approach ensures that pruned tokens do not lose their essential feature information. Discarding unimportant tokens can result in an irreversible loss of spatial and feature information, severely impacting downstream task accuracy. Our method overcomes this issue by preserving the necessary information during token compression.

Our Prune and Merge module sets itself apart from conventional token compression approaches that typically involve additional modules for calculating importance scores or merging tokens. In contrast, our module requires minimal parameters: the token mask \(\mathbf{T_M} \in \mathbb{R}^{N}\), the merge matrix \(\mathbf{M} \in \mathbb{R}^{M \times N}\), and the reconstruction matrix \(\mathbf{M^+} \in \mathbb{R}^{N \times M}\), making it highly efficient for hardware deployment. The operations involved—AND and ADD—are simple matrix operations, which ensure efficient and straightforward implementation. Specifically, the matrix multiplications in Eq.~\eqref{eq_4} and Eq.~\eqref{eq_7} are actually conducted by summing the tokens within the same merging group, weighted by their importance scores. The computational cost of both the merge and recovery operations is just \(2DN\), with the total cost of pruning and merging amounting to \(6ND\). In comparison, the computation cost of a standard ViT is \(12ND^{2} + 2N^{2}D\), which is considerably higher. As a result, the overhead introduced by our Prune and Merge module is negligible in the overall complexity calculation.

\subsection{Gradient-weighted Attention Scoring and Merge Matrix}
\label{sec3.3}
We show the detailed calculation process of our gradient-weighted attention scoring method and the generation of merge matrix and reconstruct matrix on the left part of Fig~\ref{fig_framework}. 
Our gradient-weighted attention scoring approach was inspired by Taylor pruning~\cite{molchanov2016pruning,molchanov2019importance},  which uses the first order Taylor expansions to approximate a filter’s contribution. 
For the $i$-th token $\mathbf{Z}_i$, its importance score $\mathcal I(\cdot)$ calculated by Taylor Criterion can be:
\begin{equation}
\mathcal I(\mathbf{Z}_i) = \left | \frac{1}{\mathrm{D}}\sum_{\mathrm{d}}\frac{\partial \mathcal{L}}{\partial \mathbf{Z}_{i,\mathrm{d}}}\mathbf{Z}_{i,\mathrm{d}} \right |,
\label{eq_8}
\end{equation}
where $\frac{\partial \mathcal{L}}{\partial \mathbf{Z}_{i, \mathrm{d}}}$ is the gradient of loss $\mathcal{L}$ with respect to $\mathrm{d}$-th dimension of $\mathbf{Z}_i$.
However, such calculation does not utilize the information of attention probability, which can be beneficial for learning dataset prior knowledge and is widely used for importance estimation in existing literature~\cite{tang2022patch, xu2022evo, pan2021ia, goyal2020power}. To address this, we calculate the Taylor importance of the elements on the attention map, summing it along the Query dimension to obtain the score of the corresponding token. In the case of multi-head self-attention, we average the sum of all heads. The formula is as follows:
\begin{equation}
\mathcal I(\mathbf{Z}_i) = \left | \frac{1}{\mathrm{H}} \sum_{\mathrm{h}} \sum_{j}^{\mathrm{N}}\frac{\partial \mathcal{L}}{\partial \mathbf{A}_{i, j}^\mathrm{h}}\mathbf{A}_{i, j}^\mathrm{h} \right |. 
\label{eq_9}
\end{equation}
$\mathbf{A}_{i, j}^\mathrm{h}$ denotes the $i$-th column and $j$-th row of the $\mathrm{h}$-th attention map, which represents the attention probability of the $i$-th token in Key to the $j$-th token in Query. We sum along the Query dimension to get the global attention of the $i$-th token in Key. Based on the importance scores assigned to each token, we apply a threshold, name it as PM-threshold, to obtain a token mask, where uninformative tokens are set to zero. The negation of the token mask is then used to identify and reserve the pruned tokens. Additionally, we utilize the importance scores to construct the merge matrix, where the scores of the pruned tokens are set to zero through an AND operation with the token mask.

In Algorithm~\ref{alg_1}, we elaborate on the methodology for constructing the merge matrix using importance scores. Tokens undergo classification into three distinct categories based on two specific thresholds. The first threshold, the PM-threshold, determines whether a token should be pruned or merged. Tokens that are deemed unimportant or have a negative contribution are pruned by setting their scores to zero. Tokens that are less important but would negatively impact precision if removed are merged with adjacent important tokens. The second threshold is based on the pruning rate, which helps determine whether a token should be retained as one of the most important tokens.

The generation of the merge matrix involves iterating through each row, where each row is represented as a merge vector \(\mathbf{Line}\). The elements of \(\mathbf{Line}\) are determined by the \(Left\) and \(Right\) indices of the important tokens \(\mathbf{I}\). Multiplying \(\mathbf{Line}\) with the input token sequence yields a weighted sum of the corresponding tokens, resulting in the creation of a merged token. Weights assigned to each token are determined by normalizing the scores of all tokens destined for merging. Consequently, the merge matrix is obtained, while the reconstruct matrix functions as the pseudoinverse of the merge matrix.
It is pertinent to emphasize that both the merge matrix and the reconstruct matrix are made learnable(\eg, can be optimized by loss backward), facilitating their potential effectiveness during model fine-tuning.

This methodology enables us to dynamically dictate the merging strategy based on the importance scores of tokens. This ensures the preservation of crucial information while concurrently compressing the token representation. Furthermore, by rendering the merge matrix and reconstruct matrix learnable, we introduce flexibility to adapt the compression scheme during the optimization of the model.

\begin{algorithm}[t]
	\caption{Global Pruning and Merging}
	\begin{algorithmic}[1] 
		\REQUIRE ~~\\ 
		Training dataset $\mathcal D$, vision transformer $\mathcal{T}$ with $L$ layers, iteration number $I$, $\mathcal{S}^l$ importance scores of tokens in $l$-th layer, $\mathcal{I}_{all}$ importance scores of tokens in all layers;
		\STATE Set all $\mathcal{S}_{\mathrm{sum}}^l$ to $\textbf{0}$ vector;  
		\STATE Train $\mathcal{T}$ with data sampled from $\mathcal D$;  
		\FOR{each $iter \in [1,I]$}
		\FOR{each $l\in [1,L]$}
		\STATE Calculate $\mathcal I\left ( \mathbf{Z}^l \right )$ with Eq.~\eqref{eq_9};
		\STATE $\mathcal{S}_{\mathrm{sum}}^l ( \mathbf{Z}^l ) += \mathcal{S} ( \mathbf{Z}^l )$ ;
		\ENDFOR
		\ENDFOR
		\FOR{each $l\in [1, L]$}
		\STATE $\mathcal{I}^{l} = \mathcal{I}_{\mathrm{sum}}^l / I$;
		\STATE $\mathcal{I}_{all} = \mathrm{Concat}(\mathcal{I}_{all}, \mathcal{I}^{l})$
		\ENDFOR
		\STATE Generate Merge Matrix with Algorithm~\ref{alg_1};
		\ENSURE ~~\\ 
		Compressed model; 
	\end{algorithmic}
	\label{alg_2}
\end{algorithm}

\subsection{Global Compression and Model Finetuning}
Instead of employing the handcraft structure or the searched result in a large searching space, we apply the global compression scheme to autonomously form the final structure of pruned model. Compared with hand designing and heuristic searching, global compression doesn't need much prior knowledge and extra searching time. In addition, the pruning information can be fully utilized for finding the most suitable structure of current task. 

The process of obtaining the compressed structure through global compression is illustrated in Algorithm~\ref{alg_2}. As shown in the algorithm, we calculate the importance scores of each token in each layer using an iterative averaging approach.  Subsequently, we concatenate the scores from all layers to obtain the global scores for all tokens.  After sorting the global scores, we determine the prune and reserve thresholds based on the desired number of pruned and reserved tokens. Scores below the prune threshold are set to 0, while the token indices corresponding to scores above the reserve threshold are recorded for generating the merge matrix.  Our method leverages the global impact to automatically determine the model's structure, enabling the efficient discovery of structures close to winning tickets. 

For model finetuning, we also use self-distillation loss for better accuracy recovery when finetuning. The loss function can be formulated as:
\begin{equation}
\mathcal{L} = \mathcal{L}_{CE} \left( y,p\right) + \alpha \mathcal{L}_{KL}\left( q,p\right).
\label{eq_11}
\end{equation}
$p$ is the output softmax probability of pruned model, $q$ is the corresponding value of original model, and $y$ is the true label. $\mathcal{L}_{CE}$ and $\mathcal{L}_{KL}$ denote the cross-entropy loss and the KL-divergence loss. $\alpha$ balances the proportion of the teacher's knowledge.

\begin{table*}[t]
	\renewcommand\arraystretch{1.0}
	\centering
	\caption{Comparison with state-of-the-art token pruning/merging methods on DeiT and Swin Transformer. The image resolution of all models is 224 $\times$ 224. We replicate the code of these methods for throughput tests using the same script and on the same device. $^*$ indicates the method has a different baseline.}
	\setlength{\tabcolsep}{5mm}
		\begin{tabular}{l|ccccc}
			\toprule
			\textbf{Models}                          & \textbf{Publication} & \textbf{Backbone} &  \textbf{Throughput} (img/s)  &         \textbf{FLOPs} (G)         &     \textbf{Top-1 Acc.} (\%)     \\ \hline
			Baseline~\cite{touvron2021training}      &       ICML2021       &     DeiT-Tiny     &             2234             &                1.3                 &               72.2               \bigstrut[t]\\
			SViTE~\cite{chen2021chasing}             &     NeurIPS2021      &     DeiT-Tiny     &     2498 (1.12$\times$)      &     1.0 (23.1\% $\downarrow$)      &     70.1 (2.1 $\downarrow$)      \\
			SAViT~\cite{zheng2022savit}              &     NeurIPS2022      &     DeiT-Tiny     &     2866 (1.28$\times$)      &     0.9 (30.7\% $\downarrow$)      &     70.7 (1.5 $\downarrow$)      \\
			PS-ViT~\cite{tang2022patch}              &       CVPR2022       &     DeiT-Tiny     &     3138 (1.40$\times$)      &     0.8 (38.5\% $\downarrow$)      &     72.0 (0.2 $\downarrow$)      \\
			Evo-ViT~\cite{xu2022evo}                 &       AAAI2022       &     DeiT-Tiny     &     3192 (1.43$\times$)      &     0.8 (38.5\% $\downarrow$)      &     72.0 (0.2 $\downarrow$)      \\
			DynamicViT~\cite{rao2021dynamicvit}      &     NeurIPS2021      &     DeiT-Tiny     &     3427 (1.53$\times$)      &     0.7 (46.2\% $\downarrow$)      &     71.2 (1.0 $\downarrow$)      \\
			ToMe~\cite{bolya2023token}     &       ICLR2023       &     DeiT-Tiny     &     3584 (1.60$\times$)      &     0.7 (46.2\% $\downarrow$)      &     71.3 (0.5 $\downarrow$)      \\
			PM-ViT$_{r0.7}$ (Ours)                   &          -           &     DeiT-Tiny     &     3375 (1.51$\times$)      &     0.8 (38.5\% $\downarrow$)      & \textbf{72.0 (0.2 $\downarrow$)} \\
			PM-ViT$_{r0.6}$ (Ours)                   &          -           &     DeiT-Tiny     & \textbf{3745 (1.67$\times$)} &     0.7 (46.2\% $\downarrow$)      &     71.6 (0.6 $\downarrow$)      \\ \hline
			Baseline~\cite{touvron2021training}      &       ICML2021       &    DeiT-Small     &             1153             &                4.6                 &               79.8               \bigstrut[t]\\
			SViTE(\cite{chen2021chasing})            &     NeurIPS2021      &    DeiT-Small     &     1491 (1.29$\times$)      &     3.1 (32.6\% $\downarrow$)      &     79.2 (0.6 $\downarrow$)      \\
			PS-ViT~\cite{tang2022patch}              &       CVPR2022       &    DeiT-Small     &     1614 (1.40$\times$)      &     3.0 (34.8\% $\downarrow$)      &     79.4 (0.4 $\downarrow$)      \\
			Evo-ViT~\cite{xu2022evo}                 &       AAAI2022       &    DeiT-Small     &     1623  (1.41$\times$)     &     3.0 (34.8\% $\downarrow$)      &     79.4 (0.4 $\downarrow$)      \\
			IA-RED$^2$~\cite{pan2021ia}              &     NeurIPS2021      &    DeiT-Small     &     1648  (1.43$\times$)     &     3.0 (34.8\% $\downarrow$)      &     79.1 (0.7 $\downarrow$)      \\
			EViT~\cite{liang2021evit}       &       ICLR2022       &    DeiT-Small     &     1869  (1.62$\times$)     &     2.6 (43.4\% $\downarrow$)      &     79.0 (0.8 $\downarrow$)      \\
			ToMe~\cite{bolya2023token}     &       ICLR2023       &    DeiT-Small     &     1856 (1.61$\times$)      &     2.7 (41.3\% $\downarrow$)      &     79.4 (0.6 $\downarrow$)      \\
			eTPS~\cite{wei2023joint}       &     CVPR2023      &    DeiT-Small     &     1688  (1.46$\times$)     &     3.0 (34.8\% $\downarrow$)      &     79.7 (0.1 $\downarrow$)      \\
			ToFu$_AugReg$~\cite{kim2024token}   &  WACV2024      &    DeiT-Small     &   1853  (1.61$\times$)     &     2.7 (41.3\% $\downarrow$)      &    79.6 (1.8$\downarrow$)      \\
			Zero-TP~\cite{wang2024zero} &    CVPR2024      &   DeiT-Small     &   1675  (1.45$\times$)  &    3.0 (34.8\% $\downarrow$) &   79.4 (0.4 $\downarrow$)     \\
			PM-ViT$_{r0.7}$ (Ours)                   &          -           &    DeiT-Small     &     1786 (1.55$\times$)      &     3.0 (34.8\% $\downarrow$)      & \textbf{79.8 (0.0 $\downarrow$)} \\ 
			PM-ViT$_{r0.6}$ (Ours)                   &          -           &    DeiT-Small     & \textbf{1887 (1.64$\times$)} &     2.7 (41.3\% $\downarrow$)      & 79.6 (0.2 $\downarrow$) \\ \hline
			Baseline~\cite{liu2021swin}              &       ICCV2021       &     Swin-Tiny     &             653              &                4.5                 &               81.2              \bigstrut[t] \\
			STEP-Swin-Ti~\cite{li2021differentiable} &       TACL2021       &     Swin-Tiny     &      791 (1.21$\times$)      &     3.5 (22.2\% $\downarrow$)      &      77.2 (4.0$\downarrow$)      \\
			SPViT-Swin-Ti~\cite{he2024spvit}         &      TPAMI2024       &     Swin-Tiny     &      823 (1.26$\times$)      &     3.4 (24.4\% $\downarrow$)      &      80.1 (1.1$\downarrow$)      \\
			PM-Swin$_{r0.7}$ (Ours)                  &          -           &     Swin-Tiny     & \textbf{ 876 (1.34$\times$)} & \textbf{3.1 (31.1\% $\downarrow$)} & \textbf{80.2 (1.0 $\downarrow$)} \\ \bottomrule
		\end{tabular}
	\label{tab_1}
\end{table*} 

\section{Experiments}

In this section, we evaluate the efficacy of our Prune and Merge (PM-ViT) methods in both the classification task and the challenging downstream task of semantic segmentation. To ensure a comprehensive evaluation, we carefully select state-of-the-art compression methods for comparison.

\subsection{Datasets and Metrics}

\textbf{Classification.} ImageNet-1k~\cite{deng2009imagenet} is a large-scale  dataset, consisting of 1.3 million training and 50,000 validation images with various spatial resolutions and 1000 classes. It is widely used as the benchmark for classification tasks. We report the top-1 validation accuracy and throughput of the compressed models with different compression rates, and also report Float Points Operations (denoted as FLOPs) to measure the computational complexity of models.

\textbf{Semantic segmentation.} We evaluate the downstream segmentation performance on the ADE20k~\cite{zhou2019semantic} dataset. This dataset contains intricate scenes with fine-grained labels and is one of the most challenging semantic segmentation datasets. The training set contains 20,210 images with 150 semantic classes. The validation set contains 2,000 images. We use mean Intersection-over-union (mIOU) to evaluate the instance segmentation accuracy, and measure the Frame Fer Second (FPS) to evaluate the inference speed of models.

\subsection{Token Compression for Classification}
\label{sec4.2}

\textbf{Implementation details.} We compress Deit~\cite{touvron2021training}, Swin Transformer~\cite{liu2021swin} and MAE~\cite{he2022masked} fine-tuned ViT models and compare them with the state-of-the-art token pruning/merging methods.We denote our compressed ViT model as PM-ViT and name the compressed Swin model as PM-Swin. The variable $r$ are used to represent the compression rate, and all PM-ViT models adopt a uniform PM-Threshold of 0.1, in accordance with the findings from the ablation study. We initialize the learning rate as 1e-4 and set the weight decay as 0.001. Like DeiT, we use the AdamW~\cite{loshchilov2018decoupled} optimizer and the cosine decay schedule. For compression, we first train ViT models for 500 iterations with a batch size of 256 to learn the importance score of each token and then globally prune and merge the tokens at one time. We then fine-tune the compressed model for 60 epochs to improve its performance. During finetuning, we apply self-distillation, where the $\alpha$ in Eq.~\eqref{eq_11} varies based on the model size: 0.4 for Deit-Tiny, 0.6 for Deit-Small and Swin-Tiny, and 0.8 for MAE fine-tuned ViT-Base. Additionally, we also set the merge and reconstruct matrix learnable at the first 40 epochs and freeze the matrix parameters at the last 20 epochs. Finally, we evaluate the test accuracy and throughput on a single NVIDIA Tesla V100 GPU using a default batch size of 256, unless otherwise specified.

\begin{figure*}[t]
	\includegraphics[width=0.98\linewidth]{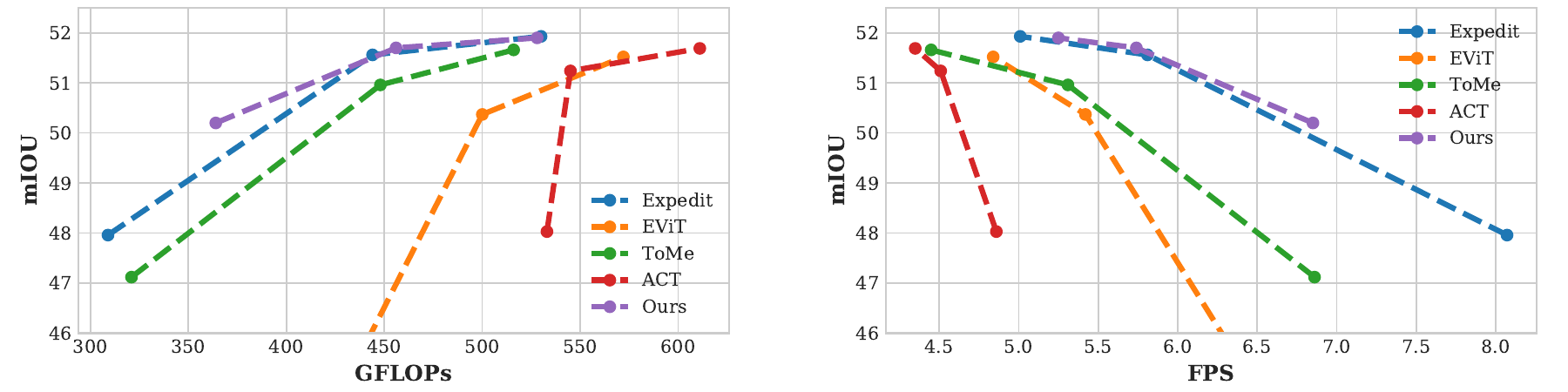}
	\vspace{-0.2cm}
	\caption{Segmenter + ViT-L compression results at various compression rates on ADE20k.  We show the comparison of our methods and state-of-the-art methods. Left: mIOU-GFLOPs curve. Right: mIOU-FPS curve.}
	\label{fig3}
\end{figure*}

\begin{figure*}[t]
	\includegraphics[width=0.98\linewidth]{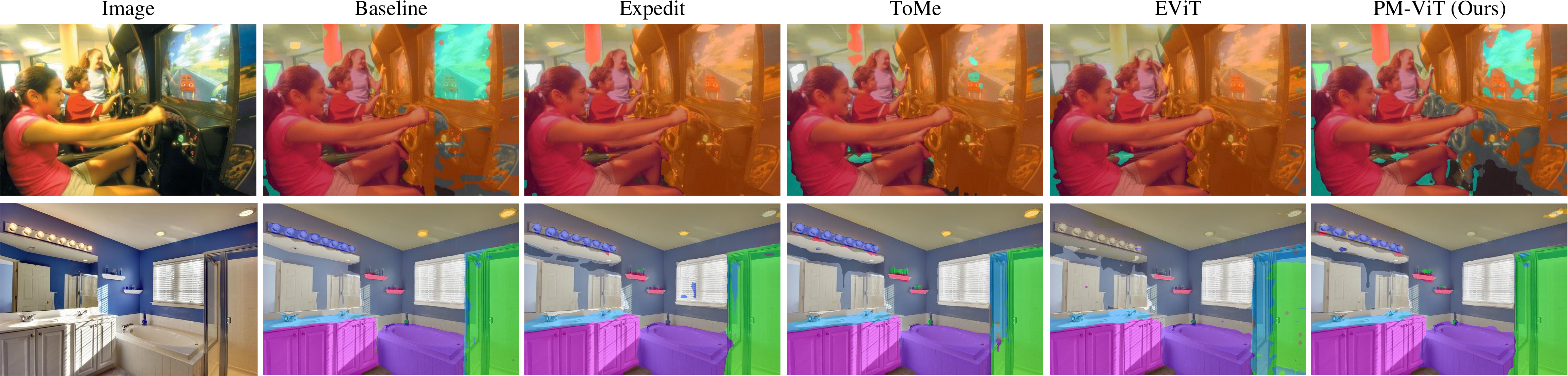}
	\vspace{-0.2cm}
	\caption{ Visualization of segmentation results on the ADE20K dataset at a 50\% token compression rate, comparing our PM-ViT with other SOTA approaches. }
	\label{fig_seg}
\end{figure*}

\textbf{Token compression on DeiTs}. Table~\ref{tab_1} compares the compression results of our Prune and Merge method with state-of-the-art methods on DeiT models. It can be observed that our compression approach achieves outstanding performance, particularly for DeiT-Tiny. With a 38.5\% reduction in FLOPs and only a 0.2\% decrease in accuracy, our PM-ViT model achieves an acceleration rate of 1.51$\times$, surpassing the performance of PS-ViT and Evo-ViT. When compared to the impressive work of ToMe~\cite{bolya2023token}, our PM-ViT with a compression rate of $r=0.6$ achieves a comparable accuracy drop and FLOPs reduction, while maintaining a higher accuracy rate. These results demonstrate the effectiveness of our Prune and Merge method, which achieves minimal accuracy degradation while providing an efficient framework for model acceleration. The experiments conducted on DeiT-Small further confirm this finding. With a similar reduction in FLOPs (34.8\%), our PM-ViT model with a compression rate of $r=0.7$ achieves the best accuracy of 79.8 and the highest acceleration rate of 1.55$\times$. Even with a more substantial token pruning (41.3\% reduction in FLOPs), our PM-ViT demonstrates minimal accuracy decline (0.2\%) and the most significant speedup (1.64$\times$). In Table~\ref{tab_3},we provide additional pruning results without fine-tuning using the AugReg~\cite{steiner2021train} method. Our PM-ViT achieves performance comparable to ToMe~\cite{bolya2023token}, with only a 0.2\% lower accuracy for both ViT-S and ViT-B. Notably, ToMe is specifically designed for pruning ViTs without training. These results indicate that our method effectively compresses tokens while efficiently accelerating the model, without compromising feature information.

\textbf{Compressed Swin Transformer.} Thanks to the flexibility afforded by our Prune and Merge method in customizing the number of compressed tokens, aligning token quantities with window attention dimensions becomes the sole requirement for ensuring compatibility with Swin Transformer compression. The compression results are illustrated in Table~\ref{tab_1}. Remarkably, most existing methods encounter challenges in adapting to the Swin model, thereby underscoring the robust generalization capability of our approach. Our PM-Swin achieves a 31.1\% reduction in the FLOPs of the Swin-Tiny model, accelerating the model by 1.34$\times$ with only a 1.0\% decrease in accuracy. While this outcome may seem less favorable compared to DieT, it primarily stems from the inherent difficulty in compressing Swin Transformer's window attention mechanism. Nonetheless, our method consistently outperforms SPViT in accuracy even at higher compression rates and acceleration ratios, thereby affirming the superiority of our approach.

\textbf{Comparison to classification models.} In Table~\ref{tab_2}, we present a comparison between our compressed MAE fine-tuned models and state-of-the-art models trained exclusively on the ImageNet-1k dataset without any additional data. Our method enables the MAE fine-tuned ViT-B model, which was initially inferior to other state-of-the-art models, to be competitive. While experiencing a slight decrease in accuracy (0.2\%), our compressed MAE fine-tuned model achieves a significant improvement in inference speed (1.67$\times$) and reduces computational complexity (by 42.1\%). Although the computational cost is higher than other models, our approach demonstrates faster inference. This finding highlights the efficiency and effectiveness of our Prune and Merge framework. 

\begin{table*}[t]
	\begin{minipage}{0.5\linewidth}
	\renewcommand\arraystretch{1.0}
	\centering
	\caption{Comparison to state-of-the-art models trained exclusively on ImageNet-1k. $^{\dag}$ indicates a different input size of 456 $\times$ 456 instead of 224 $\times$ 224. Throughputs are test on V100 with batch size of 128.} 
	\setlength{\tabcolsep}{1mm}
	\begin{tabular}{l|ccc}
		\toprule
		\textbf{Model}                                   & \textbf{Throughput (img/s)} & \textbf{FLOPs} (G) & \textbf{Top-1 Acc.} (\%) \\ \hline
		Efficient-B5$^{\dag}$~\cite{tan2019efficientnet} &            166             &        9.9         &           83.6          \bigstrut[t] \\
		Swin-S~\cite{liu2022swin}                        &            436             &        8.7         &           83.6           \\
		CSWin-S~\cite{dong2022cswin}                     &            453             &        6.9         &           83.6           \\
		MViTv2-S~\cite{li2022mvitv2}                     &            460             &        7.0         &           83.6           \\ \hline
		ViT-B$^{\mathrm{MAE}}$~\cite{he2022masked}       &            303             &        17.6        &           83.6          \bigstrut[t] \\
		\textbf{PM-ViT$_{r0.6}$ (Ours)}                  &        \textbf{505}        &        10.4        &           83.4           \\ \bottomrule
	\end{tabular}
	\label{tab_2}
	\end{minipage}
	\hfill
	\begin{minipage}{0.5\linewidth}
		
			\renewcommand\arraystretch{1.0}
			\centering
			\caption{  Pruning results without fine-tuning on the ViT-S and ViT-B models of AugReg~\cite{steiner2021train}. Throughputs are test on V100 with batch size of 128.} 
			\setlength{\tabcolsep}{1mm}
			\begin{tabular}{l|ccc}
				\toprule
				\textbf{Model}                                   & \textbf{Throughput (img/s)} & \textbf{FLOPs} (G) & \textbf{Top-1 Acc.} (\%) \\ \hline
				PS-ViT~\cite{tang2022patch}                      &            1413            &        3.6         &           79.9 (1.5 $\downarrow$)          \bigstrut[t] \\
				ToMe~\cite{bolya2023token}                        &           1427            &       3.6         &           80.9 (0.5 $\downarrow$)  \\
				PM-ViT$_{r0.8}$ (Ours)                     &            1435                  &        3.6         &           80.7 (0.7 $\downarrow$)           \\ \hline
				PS-ViT~\cite{tang2022patch}                      &            372            &        13.6         &           82.6 (2.0 $\downarrow$)          \bigstrut[t] \\
				ToMe~\cite{bolya2023token}                        &           385            &       13.6         &           83.9 (0.6 $\downarrow$)  \\
				PM-ViT$_{r0.8}$ (Ours)                     &            391                  &        13.6         &           83.7 (0.8 $\downarrow$)        \\ \bottomrule
			\end{tabular}
			
			\label{tab_3}
	\end{minipage}
\end{table*}

\subsection{Token Compression for Semantic Segmentation}

\textbf{Implementation details.} To demonstrate the effectiveness of our method on the challenging downstream task of semantic segmentation, we apply our compression technique to the ViT backbone of the Segmenter model~\cite{strudel2021segmenter}. Specifically, we compress the Segmenter + ViT-L configuration with compression rates ranging from 0.8 to 0.5 and a PM-Threshold of 0.1. For better feature learning during downstream task adaptation, we exclude the compression of the first and last two layers. The input size is set to 640 $\times$ 640. During the finetuning process, we train the model for 40 epochs using the same training set as the original Segmenter model. We employ self-distillation with a weight parameter ($\alpha$) set to 1.0 to enhance the performance. Finally, we measure the model's FPS on a single Tesla V100 GPU.

\begin{table*}[t]
	\begin{minipage}{0.5\linewidth}
		\renewcommand\arraystretch{1.0}
		\centering
		\caption{ Impact of different compression operations on model throughput. We evaluate the effect of various compression operations on model throughput by inserting these operations into the original model without compressing any tokens.}
		\setlength{\tabcolsep}{3mm}
		\begin{tabular}{l|cc}
			\toprule
			\textbf{Methods}           & \textbf{Throughput} (img/s) &  \textbf{Throughput~$\delta$}   \\ \hline
			Deit-Tiny (baseline)       &            2234             &                0                \bigstrut[t]\\ \hline
			Direct                     &            2082             &           152 (6.8\%)           \bigstrut[t]\\
			EViT~\cite{liang2021evit}  &            2183             &           115 (5.1\%)            \\
			ToMe~\cite{bolya2023token} &            2141             &           93 (4.2\%)            \\
			\textbf{PM-ViT (Ours)}     &        \textbf{2191}        & \textbf{43 (1.9\%)} \\ \bottomrule
		\end{tabular}
		
		\label{tab_4}	
	\end{minipage}
	\hfill
	\begin{minipage}{0.5\linewidth}
		\renewcommand\arraystretch{1.0}
		\centering
			\caption{Memory Usage and Computational Cost of the Prune and Merge Module (abbreviated as PM Params. and PM FLOPs, respectively) and their impact on the entire ViT model.}
			\setlength{\tabcolsep}{3mm}
			\renewcommand{\arraystretch}{1.3}
			\begin{tabular}{l|cc}
				\toprule
				\textbf{Model} & \textbf{Params. (M)} & \textbf{PM Params (M, \% of total)} \\ \hline
				ViT-Tiny & 5.7 & 0.06 (1\%) \\
				ViT-Small & 22.1 & 0.06 (0.2\%) \\ 
				\toprule
				\textbf{Model} & \textbf{FLOPs (G)} & \textbf{PM FLOPs (M, \% of total)} \\ \hline
				ViT-Tiny & 1.29 & 0.22 (0.02\%) \\
				ViT-Small & 4.68 & 0.45 (0.04\%) \\
				\bottomrule
			\end{tabular}
			
			\label{tab_complex}	
	\end{minipage}
	
\end{table*}

\textbf{Comparisons on ADE20k.} In Fig~\ref{fig3}, we compare our Prune and Merge method with state-of-the-art approaches on the ADE20k dataset, including Expedit~\cite{liang2022expediting}, EViT~\cite{liang2021evit}, ToMe~\cite{bolya2023token}, and ACT~\cite{zheng2020end}, at similar compression rates ($r=[0.8, 0.7 ,0.5]$). The visualization results in Fig.~\ref{fig_seg} further validates that our method outperforms Expedit and demonstrates a significant advantage over the other three methods. We achieve a favorable trade-off between speed and accuracy compared to the other methods. Notably, our method achieves the highest acceleration when considering equivalent computational costs.

This superior performance is attributed to the efficiency of our Prune and Merge structure. By leveraging prior information from the dataset and gradient information during training, we directly obtain the merge matrix without dynamic computation in inferencing. This significantly enhances the efficiency of the merge module. Furthermore, our combined strategy of token pruning and merging effectively preserves accuracy, ensuring competitive performance compared to other methods. Overall, our experimental results highlight the effectiveness and efficiency of our Prune and Merge method, positioning it as a promising approach for efficient model compression in resource-constrained scenarios.

\begin{figure*}[t]
	\includegraphics[width=0.98\linewidth]{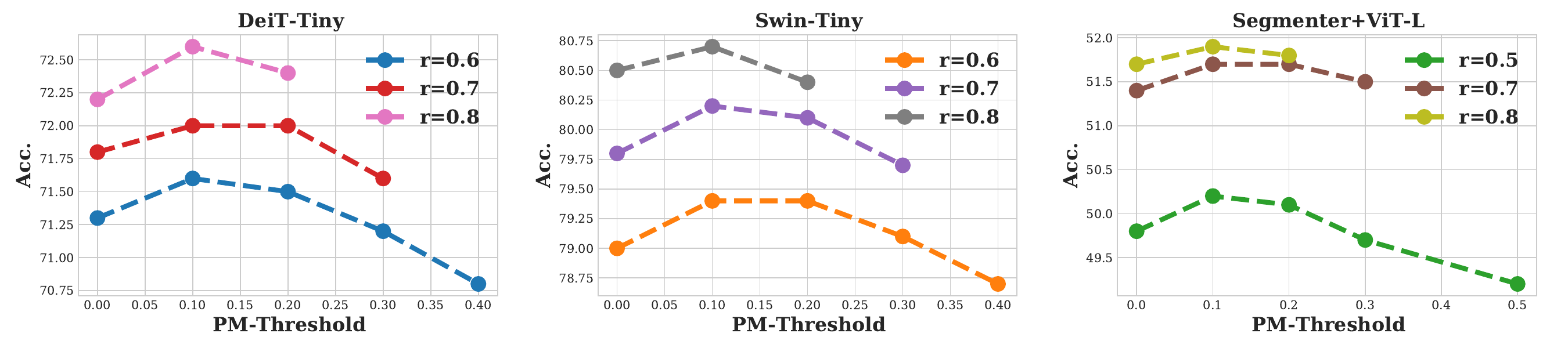}
	\vspace{-0.2cm}
	\caption{Ablation analysis of the PM-Threshold. The figure presents the accuracy versus PM-Threshold curve across different compression rates ($r$) for various models and datasets. The PM-Threshold ranges from 0 to $r$, where 0 signifies complete utilization of merging, while $r$ indicates pruning all tokens.}
	\label{fig4}
\end{figure*}

\begin{table*}[t]
	\begin{minipage}{0.4\linewidth}
		\centering
		\setlength{\tabcolsep}{1mm}
		\caption{The effect of each component in our gradient-weighted attention scoring mechanism. We report the $r=0.6$ results to reflect the compression performance. Grad and Attn are the short for gradient and attention, respectively.} 
		\begin{tabular}{l|ccc}
			\toprule
			\textbf{Methods}   & \textbf{Gradient} & \textbf{Attention} & \textbf{Top-1 Acc.} (\%) \\ \hline
			Baseline           &         -         &         -          &           72.2          \bigstrut[t] \\ \hline
			Attn only          &   \XSolidBrush    &     average token  &           69.8           \bigstrut[t]\\
			Grad  only         &    \Checkmark     &    \XSolidBrush    &           70.5           \\
			Class Attn         &   \XSolidBrush    &    class token     &           70.8           \\
			Grad \& Class Attn &    \Checkmark     &    class token     &           71.1           \\
			Random             &   \XSolidBrush    &    \XSolidBrush    &           69.3           \\
			PM-ViT (Ours) &    \Checkmark     &   average token   &      \textbf{71.6}        \\ \bottomrule
		\end{tabular}
		
		\label{tab_5}	
		
	\end{minipage}
	\hfill
	\begin{minipage}{0.28\linewidth}
		\centering
		\setlength{\tabcolsep}{1mm}
		\caption{Ablation Study on Methods for Generating Merge Matrix Weights based on Importance Scores. The term "DiV" denotes dividing the Summation, Mean, or Maximum value of the scores. }
		
		\begin{tabular}{l|c}
			\toprule
			\textbf{Methods} & \textbf{Top-1 Acc.} (\%) \\ \hline
			DiV Sum          &           70.8           \bigstrut[t]\\
			DiV Mean         &           71.1           \\
			DiV Max          &           71.4           \\
			All Ones         &           71.0           \\
			Guassian Filter  &           71.2           \\ 
			Normalize        &      \textbf{71.6}       \\ 
			Normalize + Bias &      \textbf{71.6}       \\ \bottomrule
		\end{tabular}
		
		\label{tab_6}
	\end{minipage}
	\hfill
	\begin{minipage}{0.28\linewidth}
		\centering
		\setlength{\tabcolsep}{1mm}
		\caption{The Impact of Learnable Matrix on Our Method. We conduct experiments on compressed DeiT-Tiny and DeiT-Small with various settings of the matrix. }
		
		\begin{tabular}{l|c}
			\toprule
			\textbf{Methods} & \textbf{Top-1 Acc.} (\%) \\
			\hline
			DeiT-Tiny &  72.2 \bigstrut[t]\\
			Frozen &  71.3 \\
			Learnable & 71.5 \\
			Learnable + Frozen & \textbf{71.6} \\
			\hline
			DeiT-Small &  79.8 \bigstrut[t]\\
			Frozen &  79.3 \\
			Learnable & 79.4 \\
			Learnable + Frozen & \textbf{79.6} \\
			\bottomrule
		\end{tabular}

		\label{tab_7}
	\end{minipage}
\end{table*}

\subsection{Ablation Studies}

\textbf{Impact of PM-Threshold on model accuracy.} Fig~\ref{fig4} illustrates the influence of different PM-Threshold values on model accuracy under identical compression rates. We conduct ablations on three models: DeiT-Tiny, Swin-Tiny, and Segmenter + ViT-L, across two datasets with distinct tasks: ImageNet-1k and ADE20k. It is evident that selectively pruning a small percentage of tokens (PM-Threshold 0.1 or 0.2) notably enhances model accuracy compared to employing a complete merge-based compression method (PM-Threshold 0). This enhancement can be attributed to the reduction of noise interference and improved model generalization achieved by pruning a small number of low-scoring tokens. However, as the PM-Threshold exceeds 0.3, the model's accuracy experiences a sharp decline due to the loss of important information resulting from a large number of pruned tokens. In contrast, the token merging method mitigates information loss by consolidating multiple tokens. Therefore, we advocate for a hybrid approach, involving pruning a small number of tokens (PM-Threshold 0.1) and merging others, to enhance model generalization while ensuring the effectiveness of significant compression rates.

\textbf{Efficiency of Prune and Merge.} 
Table~\ref{tab_4} showcases the impact of different token compression operations on the original model. The Direct operation involves directly utilizing the reserved tokens from the input and feeding them into the transformer block. Subsequently, the block output incorporates the pruned tokens at their original positions. This sequential operation results in a noticeable 6.8\% decrease in throughput efficiency.
The EViT has a 5.1\% impact on model throughput. It computes class attentiveness, sorts the token sequence based on their attentiveness score, and merges the less attentive tokens into a single token. It should be noted that the Argsort operation employed by EViT is not highly efficient and may not be fully supported by certain frameworks~\cite{prillo2020softsort}. Similarly, the ToMe method leverages bipartite soft matching to select tokens for merging, which also introduces inefficiency and has a 4.2\% throughput decrease.
In comparison, our Prune and Merge method exhibits a more negligible impact on model throughput, with a reduction of only 1.9\%. Importantly, our approach does not require specialized hardware support, making particularly suitable for edge device applications.

\textbf{Memory Usage and Computational Cost of Prune and Merge.} Table~\ref{tab_complex} demonstrates that our Prune and Merge module has minimal impact on both memory usage and computational cost for the ViT model. For instance, in the ViT-Tiny model, the PM Params contribute only 0.5\% of the total parameters, while the PM FLOPs account for merely 0.02\% of the total FLOPs. Similarly, in the ViT-Small model, the PM Params add only 0.1\% to the total parameters, and the PM FLOPs contribute just 0.04\% of the total FLOPs. These results highlight that the additional parameters and operations introduced by our module have a negligible impact on the overall model complexity, making our approach highly efficient and suitable for deployment.

\begin{table*}[t]
	\begin{minipage}{0.33\linewidth}
		\centering
		\setlength{\tabcolsep}{2mm}
		\renewcommand\arraystretch{1.05}
		\caption{Effect of the number of iterations for importance score calculation. The batch size for each iteration is 256.}
		\begin{tabular}{l|c}
			\toprule
			\textbf{Methods} & \textbf{Top-1 Acc.} (\%) \\ \hline
			DeiT-Tiny        &           72.2          \bigstrut[t] \\ \hline
			Random           &           70.9          \bigstrut[t] \\
			1 Iteration      &           71.1           \\
			100 Iterations    &           71.6           \\
			300 Iterations    &           71.9           \\
			500 Iterations    &         \textbf{72.0}      \\
			2000 Iterations   &          \textbf{72.0}   \\
			\bottomrule
		\end{tabular}
		
		\label{tab_8}	
		
	\end{minipage}
	\hfill
	\begin{minipage}{0.33\linewidth}
		\centering
		\setlength{\tabcolsep}{1mm}
		\caption{Robustness test to assess the impact of position augmentation on our Prune and Merge method. We incorporate random crop and resize to introduce position augmentation during fine-tuning.}
		\vspace{-0.2cm}
		\begin{tabular}{l|c}
			\toprule
			\textbf{Methods}     & \textbf{Acc. / mIOU} (\%) \\ \hline
			DeiT-Tiny            &           72.2           \bigstrut[t]\\
			PM-ViT               &           72.0           \\
			PM-ViT + Random Crop &           71.9           \\ \hline
			Segmentor            &           51.8           \bigstrut[t]\\
			PM-ViT               &           51.6           \\
			PM-ViT + Random Crop &           51.6           \\ \bottomrule
		\end{tabular}
		
		\label{tab_9}
	\end{minipage}
	\hfill
	\begin{minipage}{0.33\linewidth}
		\centering
		\setlength{\tabcolsep}{1mm}
		\caption{Effect of self-knowledge distillation on accuracy during finetuning for three models. No kd indicates finetuning without knowledge distillation.}
		
		\begin{tabular}{l|c}
			\toprule
			\textbf{Methods}     & \textbf{Acc. / mIOU} (\%) \\ \hline
			PM-ViT (no kd)       &           71.7           \bigstrut[t]\\
			PM-ViT               &           72.0           \\ \hline
			PM-Swin (no kd)      &           79.7           \bigstrut[t]\\
			PM-Swin              &           80.2           \\ \hline
			PM-Segmenter (no kd) &           51.3           \bigstrut[t]\\
			PM-Segmenter         &           51.6           \\ \bottomrule
		\end{tabular}
		\label{tab_10}
	\end{minipage}
\end{table*}

\textbf{Effectiveness of attention and gradient guidance.} We examine the impact of different components in importance calculation, as shown in Table~\ref{tab_5}. The results reveal that the absence of gradient information leads to a significant decline in performance for compressed models. This aligns with our argument that gradient information reflects the global impact on the model. Conversely, solely utilizing gradient information without the guidance of attention maps also results in a noticeable decrease in accuracy, although the impact is comparatively smaller than when gradient information is absent.

Additionally, we explore the use of different token attentions, specifically the class token and the average of all tokens. Our findings indicate that the class token exhibits a higher value as it provides clearer global attention information. However, when gradients are weighted, the attention results demonstrate that the average token outperforms the class token. We contend that the average token incorporates more gradient information in the calculation, compensating for its relative deficiency when compared to the class token.   

\begin{figure*}[t]
	\includegraphics[width=0.98\linewidth]{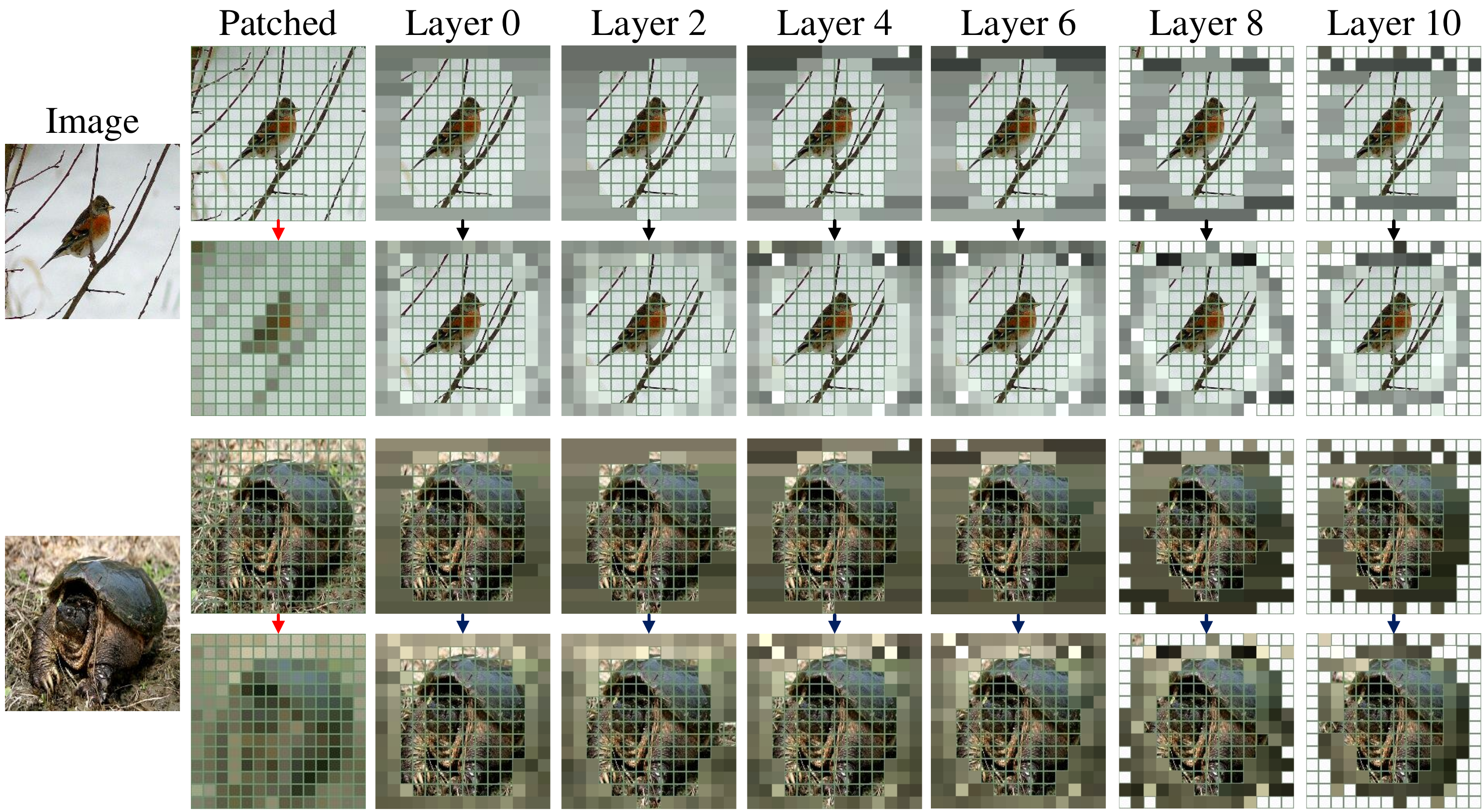}
	\caption{ Visual effects of token pruning and merging on ImageNet-1k validation set. The red arrow denotes patches tokenized into corresponding tokens, while the black arrow represents the recovery of merged tokens using the reconstruct matrix. Reserved tokens are displayed with patches instead of individual tokens to ensure clarity. Pruned tokens are depicted as white, while merged tokens are visualized based on their merging results.}
	\label{fig5}
\end{figure*}

\textbf{Ablation study on matrix generation methods.} We investigate various mapping techniques to derive merge matrix parameters based on token importance scores. The results are presented in Table~\ref{tab_6}. Initially, we compute the sum, mean, and maximum of these scores and normalize them for mapping. However, these methods exhibit unsatisfactory results, with an accuracy drop ranging from 1.4\% to 0.8\%. We also explore alternative approaches, such as setting parameters to all ones or applying Gaussian filtering to the scores, but these methods prove ineffective, resulting in an accuracy drop of 1.2\% and 1\%, respectively. Notably, the most effective approach is to normalize the scores and utilize them as weight parameters. Furthermore, ToMe~\cite{bolya2023token} proposes incorporating bias before softmax calculation for merged tokens to mitigate the impact on the softmax attention outcome. However, our attempts to integrate this approach do not result in any performance improvement, regardless of using learnable or frozen bias. We hypothesize that our normalization method has minimal impact on the softmax attention outcome, rendering the bias for token size adjustment unnecessary. Considering the model's efficiency, we decide not to include this mechanism. 

\textbf{Effect of learnable matrix.} Table~\ref{tab_7} presents the results obtained by applying different fine-tuning schedules to the same compressed models. Directly using the frozen initial parameters of the merge matrix and reconstruct matrix yields unsatisfactory results, with a drop in accuracy of 0.9\% and 0.5\%, respectively. By making the matrix trainable, we enable the optimization of its weights during fine-tuning, thereby enhancing the effectiveness of our Prune and Merge method, resulting in an increase of 0.2\% and 0.1\% compared to the frozen parameters.
However, we find that the improvement achieved by optimizing the matrix weights is limited in the fine-tuning process. We consider that the model parameters cannot be optimized to their best with a changed matrix. To address this, we freeze the matrix weights during the last 5 epochs of fine-tuning and optimize the model weights based on the fixed matrix parameters. This approach further improves the performance of our method, with an additional increase of 0.1\% and 0.2\%.
Overall, we decide to choose the learnable + frozen schedule for our fine-tuning.

\textbf{Impact of iteration numbers on our method.} Table~\ref{tab_8} showcases the performance of our method as the number of iterations varies, compared with random compression.  Our approach improves inference efficiency by leveraging the prior distribution of the dataset. With only 1 iteration, our method performs closely to random compression.  As the number of iterations increases, the performance of our method demonstrates a noticeable improvement as the result of learning dataset prior knowledge.  However, beyond 500 iterations, the performance improvement becomes marginal, indicating that our approach has sufficiently captured the prior information of the dataset.

\textbf{Robustness test on the influence of target position.} Our method enhances the Prune and Merge strategy by incorporating prior knowledge from the training dataset, showing high efficacy in classification tasks with stable target positions. However, it faces challenges with random transformations and in semantic segmentation, where target positions vary. We present the results in Table~\ref{tab_9}. Despite a 0.2 dip in accuracy due to random cropping in the DeiT-Tiny model, our method exhibits notable robustness. The slight decrease suggests that the learnable matrix effectively adapts to positional changes during fine-tuning. We conducted similar experiments on semantic segmentation datasets, and the mIOU remained unchanged. Despite this, we believe that there are still effects. We hypothesize that introducing additional position augmentation may enhance performance, thereby compensating for the losses incurred by these effects.

\textbf{Effect of self-knowledge distillation.} We show the results of our PM-ViT without self-knowledge distillation on Table~\ref{tab_10}. We can observe that our self-knowledge distillation method typically enhances accuracy/mIOU by approximately 0.4\%. This improvement is commonly observed, as many pruning/merging methods utilize similar self-knowledge distillation techniques.

\textbf{Visualization analysis.}  
In Fig.~\ref{fig5}, we provide a visual representation of the impact of our Prune and Merge method on image patches. By obtaining the merge matrix for each layer and applying it to the token sequence, we showcase the effects of our pruning and merging operations. To enhance clarity, the reserved tokens are replaced with patches in the visualization. In order to preserve the spatial structure of the original image, merged patches are depicted as the result of the merging process, while pruned patches are shown as white. Furthermore, we multiply the merged tokens with the reconstruct matrix to demonstrate the recovery results of our methods.

From the visualization, it is evident that patches containing objects of interest in the image are deemed highly important and are neither pruned nor merged. Conversely, regions surrounding the target are considered less important, resulting in their merging with neighboring patches. The edge regions of the image, being the least important, undergo pruning with numerous patches being removed. Notably, the reconstructed tokens exhibit a similar distribution to the tokens prior to merging, thus validating the efficacy of our design in preserving token features and spatial information.

Furthermore, the specific pruning and merging schemes vary across layers. As the depth of the layers increases, the compression rate of tokens becomes higher. Specially, pruned tokens are predominantly concentrated in the last few layers, where deep-layer tokens exhibit higher similarity and redundancy in Transformers, aligning with the observation in~\cite{zong2021self}.
Our proposed Prune and Merge structure demonstrates an efficient approach for achieving layer-wise token pruning and merging. By leveraging this approach, we effectively enhance the compression results of tokens.

\section{Conclusions}
\label{conclusion}
In this work, we introduce the Prune and Merge (PM-ViT) method, an efficient token compression approach for vision transformers. Our method seamlessly combines token pruning and merging within the Prune and Merge module, effectively reducing the overall token count while mitigating information loss. The module achieves layer-wise token compression and requires only minimal parameters, making it highly suitable for hardware deployment.
To generate the module parameters, we propose a gradient-weighted attention scoring mechanism, eliminating the need for separate token importance calculations during inference. Additionally, we present an automatic global token compression approach that identifies winning tickets using gradient information.
Extensive experimental evaluations on diverse datasets and downstream tasks validate the effectiveness and versatility of PM-ViT, surpassing state-of-the-art methods in accuracy, model size, and inference speed. Visualizations and ablation studies provide valuable insights into its behavior.
Our work significantly contributes to the advancement of model compression techniques, paving the way for further research in this field. In the future, we will explore methods that leverage inference-time attention information while maintaining efficient layer-wise compression.

\bibliographystyle{IEEEtran}
\bibliography{egbib}

\end{document}